\documentclass{llncs}
\usepackage{llncsdoc}
\usepackage{times}
\usepackage{epsfig}
\usepackage{graphicx}
\usepackage{amsmath}
\usepackage{amssymb}
\usepackage{algorithm} 
\usepackage{algorithmicx,algpseudocode}
\usepackage{float}
\usepackage{comment}
\usepackage[margin=1.3in]{geometry}
\usepackage{subfig}

\begin{document}
\newpage
\title{Deep Recurrent Level Set for Segmenting Brain Tumors}

\author{T. Hoang Ngan Le \hspace{1cm} Raajitha Gummadi \hspace{1cm} Marios Savvides\\
{\tt\small {thihoanl, rgummadi}@andrew.cmu.edu}
\hspace{1cm}
{\tt\small msavvid@ri.cmu.edu}
}
\institute{Carnegie Mellon University}
\maketitle
% \tableofcontents
% \newpage
\vspace{-2em}
\begin{abstract}
Variational Level Set (VLS) has been a widely used method in medical segmentation. However, segmentation accuracy in the VLS method dramatically decreases when dealing with intervening factors such as lighting, shadows, colors, etc. Additionally, results are quite sensitive to initial settings and are highly dependent on the number of iterations. In order to address these limitations, the proposed method incorporates VLS into deep learning by defining a novel end-to-end trainable model called as Deep Recurrent Level Set (DRLS). The proposed DRLS consists of three layers, i.e, Convolutional layers, Deconvolutional layers with skip connections and LevelSet layers. Brain tumor segmentation is taken as an instant to illustrate the  performance of the proposed DRLS. Convolutional layer learns visual representation of brain tumor at different scales. Since brain tumors occupy a small portion of the image, deconvolutional layers are designed with skip connections to obtain a high quality feature map. Level-Set Layer drives the contour towards the brain tumor. In each step, the Convolutional Layer is fed with the LevelSet map to obtain a brain tumor feature map. This in turn serves as input for the LevelSet layer in the next step. The experimental results have been obtained on BRATS2013, BRATS2015 and BRATS2017 datasets. The proposed DRLS model improves both computational time and segmentation accuracy when compared to the the classic VLS-based method. Additionally, a fully end-to-end system DRLS achieves state-of-the-art segmentation on brain tumors.
%    Variational Level Set (VLS) has been a widely used method in medical  segmentation. However, the segmentation accuracy in the VLS methods dramatically drops down when dealing with numerous intervening factors such as lighting, shadows, colors, etc. In addition, its segmentation results are quite sensitive to initial settings and highly depend on the number of iterations. To address these limitations, we incorporate the VLS into a learnable deep learning by defining a new model Deep Recurrent Level Set (DRLS). For brain tumor segmentation, the local, global, and contextual information of magnetic resonance imaging (MRI) scans is important to obtain a high quality feature map. To achieve this goal, the proposed DRLS model contains three main type of layers: convolutional layers, deconvolutional layers and LevelSet layers. In each step, we feed forward both the input feature map and a Level Set map to obtain the predicted Level Set map which in turn serves as the input for the next step. The experimental results have shown that our proposed RDLS improves both computational time and segmentation accuracy against the classic VLS-based method whereas the fully end-to-end system DRLS achieves the state-of-the-art segmentation on brain tumors.
  
\end{abstract}

%%%%%%%%% BODY TEXT 
\section{Introduction}
\label{sec:introduction}

%According to CBTRUS (Central Brain Tumor Registry of the United States), an estimated 78,980 new cases of primary malignant and non-malignant brain and other CNS tumors are expected to be diagnosed in the United States in 2018. Moreover, 16,616 deaths are likely to be attributed to primary malignant brain and other CNS tumors in the US in 2018. Therefore, there is a need for efficient, early detection and diagnosis of brain tumors. 

%The Human brain consists of three regions: white matter (WM), grey matter (GM) and cerebrospinal fluid (CSF). A brain tumor is the abnormal growth of cells within the brain or the central spinal canal. It maybe cancerous or benign depending on the grade of tumor namely, gliomas and meningiomas respectively. Since Gliomas are cancerous in nature, they are of more interest in the early detection of cancer. Gliomas are tumors of glial cells in the brain. They are infiltrative in nature, most commonly occur near the white matter fibre and have ill defined boundaries. However, they can spread to any part of the brain making them very difficult to detect and segment them accurately. Gliomas tumors are generally divided into four grades grouped into two categories, namely, low grade gliomas (LGG - grade one and grad two) and high grade gliomas (HGG - grade three and grad four). 

According to CBTRUS (Central Brain Tumor Registry of the United States), an estimated 78,980 new cases of primary malignant and non-malignant brain and other CNS tumors are expected to be diagnosed in the United States in 2018. Moreover, 16,616 deaths are likely to be attributed to primary malignant brain and other CNS tumors in the US in 2018. Magnetic resonance imaging (MRI) and computed tomography (CT) scans provide high-resolution images of the brain. Based on the degree of excitation and repetition times, different modalities of MRI images maybe obtained, i.e. Fluid attenuation inversion recovery (FLAIR), spin-lattice relaxation T1-weighted (also referred to as T1), spin-spin relaxation T2- weighted (also referred to as T2) and T1-weighted contrast-enhanced (gadolinium-DTPA) also referred to as T1C. These modalities prove to be highly useful in detecting different subregions of the brain tumor, namely: edema (Whole tumor), non-enhancing solid core (tumor core), necrotic/cystic core and enhancing core.
%while being non-invasive unlike their earlier counterparts such as pneumo-encephalography and cerebral angiography. 
%For the most part, MRI is chosen for evaluating patients who have symptoms suggesting the presence a brain tumor. 
Manual detection, segmentation of the brain tumors for cancer diagnosis, from large amounts of MRI images generated during clinical routine, is a difficult and time consuming task. Thus, there is substantial importance for automatic brain tumor image segmentation from the magnetic resonance imaging (MRI) for diagnosis and radiotherapy. 

A fundamental difficulty with segmenting brain tumors automatically is that they can appear anywhere in the brain, and vary in their shape, size and structure.Additionally, Brain tumors along with their surrounding edema are often diffused, poorly contrasted, and have extended tentacle-like structures. 
VLS method with Active Contouring (AC) is widely applied in image segmentation \cite{osher1988fronts} due to its ability to automatically handle such various topological changes. 
%LS methods starts with a guesstimate contour defined by an implicit function in a higher dimension to represent contours, known as zero level set. The zero level sets are iteratively updated by minimizing the energy between inside and outside the contour. The energy function forces the contour to evolve such that it accurately segments foreground from the background when it reaches a minimum. The value of a pixel which is used as the segmentation label is defined as the signed distance between the pixel and the segmentation boundary. Thus pixels inside the contour take a positive value while pixels outside the contour take negative values.Final segmentation can be computed by a Heaviside transformation that projects negative values to 0 (outside) and positive values to 1 (inside). The contour is evolved according to a partial differential equation (PDE) derived from a Lagrangian formulation of AC model. 
Some of the remarkable works \cite{15}, \cite{16}, \cite{17} on brain tumor segmentation that utilized VLS have shown the potential of VLS in achieving highly accurate brain tumor segmentation. However, the segmentation accuracy in the VLS based methods dramatically reduces when dealing with numerous intervening factors such as lighting, shadows, colors and backgrounds with large variety or complexity. MRI images are modalities that contain such factors. The limitations of the VLS approaches can be surmised as follows: Firstly, VLS methods are largely handicapped in capturing variations of real-world objects due to its sole dependency on pixel values. Secondly, VLS methods fail to memorize and to fully infer target objects since they do not have any learning capability. Thirdly, VLS based methods are limited in segmenting multiple objects with semantic information. Finally, the segmentations generated by the LS methods are quite sensitive to numerous predefined parameters such as initial contour and number of iterations.o````1''''
%Level sets provide accurate segmentations while Deep learning algorithms show the ability to learn parameters. This sparked the most recent developments that combined Level Sets with Deep learning. \cite{18} \cite{19} \cite{20} proposed a novel image segmentation approach that integrates fully convolutional networks (FCNs) with a level set model \cite{21}. However, these methods did not apply their mo-dels to brain tumor segmentation.
%In the following sections, existing methods in brain tumor segmentation is breifly explored. This is followed by a description of the related concepts to the proposed approach. Consequently the proposed approach is explained and evaluated.
To overcome the limitation of VLS in solving the problem of brain tumor segmentation, our motivation is to answer the following questions: (1) VLS provides accurate segmentations but depends on parameters while deep neural network algorithms show the ability to learn parameters. Thus the question lies as to \textit{ How to incorporate LS into deep learning to inherit the merits of both LS algorithm and deep learning?} (2) most deep learning-based semantic image segmentation methods perform the segmentation task using a softmax function. \textit{Is it possible to replace the softmax function by a LS energy minimization to get a better outcome on MRI images? If so, how is curve evolution performed with forward and backward processes of the deep learning framework}. (3) In LS-based approach, the foreground background separation depends on the zero LS function which is computed by a sequence of iterations.\textit{ How are such iteration processes performed in the deep framework?}
 
 \begin{comment}
\begin{itemize}
\item VLS provides accurate segmentations but depends on parameters while deep neural network algorithms show the ability to learn parameters. Thus the question lies as to \textit{ How to incorporate LS into deep learning to inherit the merits of both LS algorithm and deep learning?}
\item Most deep learning-based semantic image segmentation methods perform the segmentation task using a softmax function. \textit{Is it possible to replace the softmax function by a LS energy minimization to get a better outcome on MRI images? If so, how is curve evolution performed with forward and backward processes of the deep learning framework}
\item In LS-based approach, the foreground background separation depends on the zero LS function which is computed by a sequence of iterations.\textit{ How are such iteration processes performed in the deep framework?}
\end{itemize}
\end{comment}
To address these issues and boost the classic VLS methods to learn-able deep network approaches, we propose a new formulation of VLS integrated in a deep framework, called the \textbf{Deep Recurrent Level Set (DRLS)} which combines the advantages of both fully convolutional network  (FCN)\cite{long2015fully} and LS method \cite{chan2001active}. For MRI images, the local, global, and contextual information is important to obtain a high quality feature map. To achieve this goal, the proposed DRLS is designed by incorporating LevelSet Layers into VGGNet-16 \cite{simonyan2014very} with three types of layers: Convolutional layers, deconvolutional layers and LevelSet layers. The proposed DRLS contains two main parts corresponding to visual representation and curve evolution. The first part extracts features using a Fully Convolutional Networkk (FCN) while incorporating skip connection to up-sample the feature map. The second part is composed of a level set layer that drives the contour such that the energy function attains a minima as shown in Fig.\ref{fig:flowchart}. Notably, our target is to show that it is completely possible to promote VLS to the higher level of learnable framework.

%The proposed DRLS combines the advantages of both fully convolutional network  (FCN)\cite{long2015fully} and LS method \cite{chan2001active} by incorporating the recurrent connections into each convolutional layer. In the $t^{th}$ step, we feed forward the LevelSet  feature map ($\textbf{x}$) extracted from the input image ($\textbf{I}$) through the LevelSet layer to obtain the \textit{predicted LevelSet} which in turn serves as the LS map ($\phi_{t+1}$) in the next step $(t+1)^{th}$. Figure \ref{fig:flowchart} demonstrates the architecture overview of the proposed DRLS model. The proposed DRLS contains two main parts corresponding to visual representation and curve evolution. The first part extracts features using a Fully Convolutional Networkk (FCN) while incorporating skip connection to up-sample the feature map. The second part is composed of a level set layer that drives the contour such that the energy function attains a minima.
\section{Literature review}
Over the years, discriminative, generative and deep learning methods have been used to segment brain tumors from MRI images. The following is brief description of such methodologies.

\textbf{Classical segmentation methods}
Discriminative methods mostly use supervised learning techniques to learn the relationship between the input image and the ground truth by learning the features. 
%Thus, they require large data sets with valid ground truth
Anitha et al \cite{1}, proposed segmentation using adaptive pillar K-means followed by extracting crucial features from the segmented image using discrete wavelet transforms. The features are put through two-tier classifiers namely, k-Nearest Neighbor Classifier(k-NN) and self-organizing maps(SOM). 
%The algorithm showed improved accuracy and specificity when compared with SVM based classification. 
Dimah et al \cite{2} proposed a level set based approach for tumor segmentation by using histogram based clustering. The method also provides a local statistical characterization of the image by integrating the probabilistic non-negative matrix factorization (PNMF) framework into level set formulation. The model showed more robustness to noise and intensity inhomogeneity. Few papers make use of Support Vector Machines such as in \cite{3} and \cite{4}. \cite{4} incorporates Conditional random fields to refine the segmentation. Tustison et al \cite{5} used asymmetry and first order statistical features to train concatenated Random Forests (RF) by introducing the output of the first RF as an input to the another. 
%The method achieved a Dice overlap measure of 87\%, 78\%, and 74\% for whole tumor, core, and active tumor regions, respectively in MICCAI BRATS 2013 challenge. As an extension Ellwaa et al \cite{6} made use of Random Forest with 45 trees iteratively on BRATS dataset.  

\textbf{Level-Set Methods}
Some of the initial works that utilized Level-Sets for brain tumor segmentations are \cite{15} and \cite{16}. \cite{15} combined Level set evolution and global smoothness with the flexibility of topology changes followed by mathematical morphology. Thus achieving significant advantages over conventional statistical classification. The method evaluated the working of the algorithm based on volume overlap and Haussdorf distance. A major challenge of level set algorithms is to set the equation parameters: more specifically the speed function. \cite{16} introduced a threshold-based scheme that uses level sets for 3D tumor segmentation (TLS). A global threshold is used to design the speed function which is defined based on confidence interval and is iteratively updated(search-based and adaptive) throughout the evolution process that require different degrees of user involvement. Thapaliya et al\cite{17} introduced a new signed pressure function (SPF) that can efficiently stop the contours at weak or blurred edges. The algorithm differentiates tumors from the rest of the image using local statistics. Additionally, calculations of basic thresholding and therefore different parameters for different types of images were automatic.

%Although, the traditional classification methods report high performances, the ability to learn highly discriminative features often outperforms hand-crafted and pre-defined feature sets. Thus, the modern trend of brain tumor segmentation techniques make extensive use of deep learning methods and show state-of-the-art results. 

\textbf{Deep Learning Methods}
%Ciresan et al \cite{7}. (2012) presented GPU implementation of a two-dimensional CNN for the segmentation of neural membranes.
In the year 2015, the top finisher of BRATS 2015 challenge was the first to apply Convolution Neural Networks (CNN) to brain tumor segmentation \cite{8}. The proposed CNN architecture exploits both local features as well as more global contextual features simultaneously and was 30 times faster than the then state of art solutions. Additionally, the architecture uses convolutional implementation of a fully connected layer thereby allowing a 40 fold speed up. 
%\cite{8} implemented a cascaded two-pathway CNN architecture . The CNN processes local details using smaller sized patches of 33x33 and larger context of brain tissue using patches of size 65x65 extracted at the same time. 
Urban et al \cite{9}. proposed a 3D CNN architecture which extracts 3D voxel patches from different brain MRI modalities.
%for the multi-modal MRI glioma segmentation task on BRATS dataset. The method extracts 3D voxel patches from different brain MRI modalities. 
The tissue label of the center voxel is predicted by feeding 3D voxels into a 4-layered CNN architecture. 
%Input layer consists of fifteen 3D filters followed by 2 hidden layers of 25 units each and a 6 unit softmax layer. 
%The method showed  dice scores of 87\% for the whole tumor region, 77\% for the core tumor region and 73\% for the active tumor region. 
%Thus we can surmise that CNNs are also capable of easily handling 4D data. 
In order to avoid high computations of 3D voxels, Zikic et al \cite{10} transformed the 4D data into 2D data such that standard 2D-CNN architectures can be used to solve the brain tumor segmentation task

%i.e 4-modality 3D patches of dimensions say $ (d_1*d_2*d_3*4) $ into $ (d_1*d_2*4d_3) $, 
%such that standard 2D-CNN architectures can be used to solve the brain tumor segmentation task. 
%The method reported dice scores of 83.7\% for the whole tumor region, 73.6\% for core tumor region and 69\% for active tumor region on limited BRATS Dataset.

%The model showed high BRATS dice scores of 88\% for whole tumor region, 79\% for core tumor region and 73\% for active tumor region.
Recently, \cite{11} evaluated a 11-layered CNN architecture on BRATS dataset by implementing small 3 x 3 sized filters in the convolutional layers and reported comparative dice scores.
%The method reported  dice scores of 88\%, 83\% and 77\% for whole tumor, core tumor and active tumor regionsrespectively. 
%Thus, more convolutional layers can be added to the architecture without reducing the effective receptive field of the traditional bigger filters.
In order to improve the performance and overcome the limitation of training data, CNNs is designed in a another fashion which combines with other classification methods or clustering methods \cite{12}
%In \cite{12}, patches of labels are extracted from ground truth images and then clustered by k-means algorithm into N groups to form a label patch dictionary of size N
%One such method combines local structure prediction with CNN \cite{12}. Patches of labels are extracted from ground truth images and then clustered by k-means algorithm into N groups to form a label patch dictionary of size N. 
%This is followed by a 2D CNN to classify multimodal input image patches into one of the N clusters. 
%The technique reported dice scores of 83\%, 75\% and 77\% for whole tumor, core tumor and active tumor regions respectively on BRATS dataset.
%In recent 2016 BRATS challenge, Chang \cite{13} proposed a network with a dense output classification matrix. The architecture combines activations from the deepest convolutional output layer with hyperlocal features from the original input image just prior to the final segmentation. 
%The method reported BRATS dice scores of 87\%, 81\% and 72\% for whole tumor, core tumor and active tumor regions respectively. 
One of the state of the art deep learning - based approach for segmenting brain tumor was developed by \cite{14} called DMRes which is an improvement of Deep Medic \cite{KAMNITSAS201761}. 

%Deep Medic with Residual connections (DMRes) proposed in 2016 BRATS challenge by Kamnitsas et al \cite{14}, was improved from its previous version, Deep Medic \cite{KAMNITSAS201761}, a 11-layer deep, double-pathway, 3D Convolutional Neural Network by adding residual connections between the outputs of every two layers, except for the first two of each pathway to enforce abstracting away from raw intensity values. 

%DMRes reported dice scores of 89.6\% 76.3\% and 72.4\% for  whole tumor, core tumor and active tumor regions respectively. This is an improvement over Deep Medic's dice scores of 89.6\% 75.4\% and 71.8\% for  whole tumor, core tumor and active tumor regions respectively.

%Although there have been significant contributions to the field of brain tumor segmentation.  No significant work or experimentation has been reported in which deep learning and level sets were integrated in an algorithm as proposed.
\section{Proposed Network}
\begin{figure}[H]
    \centering
    \subfloat[The proposed DRLS network with two main parts: visual representation by recurrent FCN and curve evolution by the proposed LevelSet layer]{{\includegraphics[width=0.62\columnwidth]{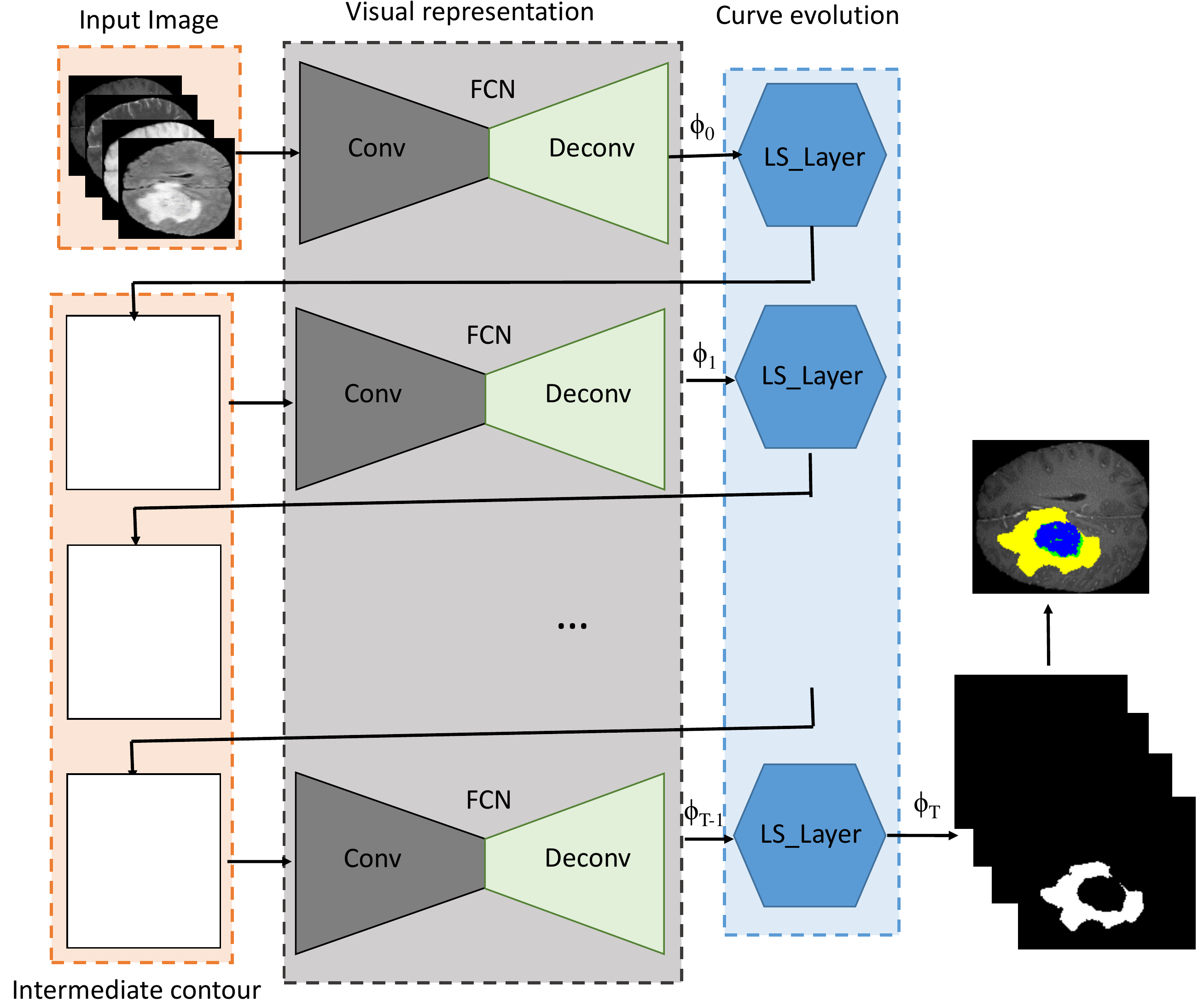} }}%
    \qquad
    \subfloat[Psedo code of the proposed network]{{\includegraphics[width=0.28\columnwidth]{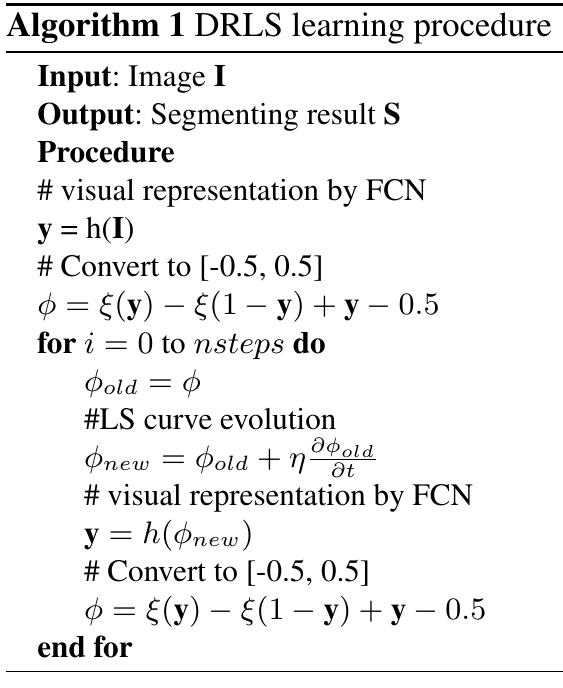} }}%
    \caption{The proposed DRLS network and algorithm}%
    \label{fig:flowchart}%
\end{figure}

The pipeline of the proposed network is as illustrated in Figure. \ref{fig:flowchart}. In this section, a review of formulation of the traditional LS is provided. Then, the proposed Recurrent Fully Convolutional Neural Network (RFCN) is introduced. Finally, the proposed end-to-end Deep Recurrent Level Set (DRLS) which incorporates VLS into RFCN is detailed.
\subsection{Formulation of Level Sets}
\label{sec:LSformular}
Consider a binary image segmentation problem in 2D space, $\Omega$. The boundary $C$ of an open set $\omega \in \Omega$ is defined as: $C = \partial \omega$. In VLS framework, the boundary $C$ can be represented by the zero level set $\phi$ as follows:
\begin{equation}
\forall (x,y) \in \Omega
\left\{\begin{matrix}
C =  &  \{(x,y) : \phi(x,y) = 0\}\\ 
inside (C)) &  \{(x,y) : \phi(x,y) > 0\}\\ 
output (C)) &  \{(x,y) : \phi(x,y) < 0\} 
\end{matrix}\right.
\end{equation}

For image segmentation, $\Omega$ denotes the entire domain of an image $\textbf{I}$. The  zero LS function $\phi$ divides the region $\Omega$ into two regions: region inside $\omega$ (foreground), denoted as inside(C) and region outside $\omega$ (background) denoted as outside(C). The length of the contour C is defined as: $Length(C)  =  \int_{\Omega} {|\nabla H(\phi(x,y))|dxdy} = \int_{\Omega} {\delta(\phi(x,y))|\nabla \phi(x,y)| dxdy}$ and the area inside the contour C is defined as $Area(C) = \int_{\Omega} H(\phi(x,y))dxdy$ 
\begin{comment}
\begin{eqnarray}
\begin{split}
&Length(C) & = & \int_{\Omega} {|\nabla H(\phi(x,y))|dxdy} = \int_{\Omega} {\delta(\phi(x,y))|\nabla \phi(x,y)| dxdy}\\
& Area(C) & = &\int_{\Omega} H(\phi(x,y))dxdy
\end{split}
\end{eqnarray}
where, $H_{\epsilon} ({\cdot})$ is a Heaviside function $H_{\epsilon} ({\tau}) = \dfrac{1}{2} \left( 1+\frac{2}{\pi}\tan^{-1} \left( \frac{{\tau}}{\epsilon}\right) \right) $. The Dirac delta derivative of Heaviside is $\delta _{\epsilon} ({\cdot}) = H'_{\epsilon}(\cdot)  = \frac{\epsilon}{\pi (\epsilon ^2+{\tau}^2)}$
%. They are defined as:
\begin{eqnarray} 
%\small
H_{\epsilon} ({\tau}) = \dfrac{1}{2} \left( 1+\frac{2}{\pi}\tan^{-1} \left( \frac{{\tau}}{\epsilon}\right) \right) \\
\delta _{\epsilon} ({\tau})=\frac{\partial}{\partial \tau}H_{\epsilon} ({\tau}) = \frac{\epsilon}{\pi (\epsilon ^2+{\tau}^2)}
\label{eq:Heaviside_Delta}
\end{eqnarray}
\end{comment} 

Typically, the LS-based segmentation methods start with an initial level set $\phi_0$ and an given image $\textbf{I}$. The LS updating process is performed via gradient descent by minimizing an energy function which defined based on the difference of image features, such as color and texture, between
foreground and background. LS utilizes shape and regions to improve the performance. Since LS uses only low-level features, it is limited when reading complex images. However, to compensate this limitation, deep networks have the ability to learn and encode useful high-level features.

\subsection{Recurrent Fully Convolutional Neural Network}

CNNs contain a set of building blocks of Neural Networks (NNs) which have shared weights across different spatial locations and are based on translation invariance with three main components, namely, convolution, pooling, and activation functions. The output feature map is obtained by convolving convolution kernels with the input feature map of fixed size. 

\begin{figure}[!t]
    \centering
    \subfloat[Comparison of different deep models. (1) Convolution network. (2) Fully convolutional network. (3): recurrent fully convolutional network]{{\includegraphics[width=0.35\columnwidth]{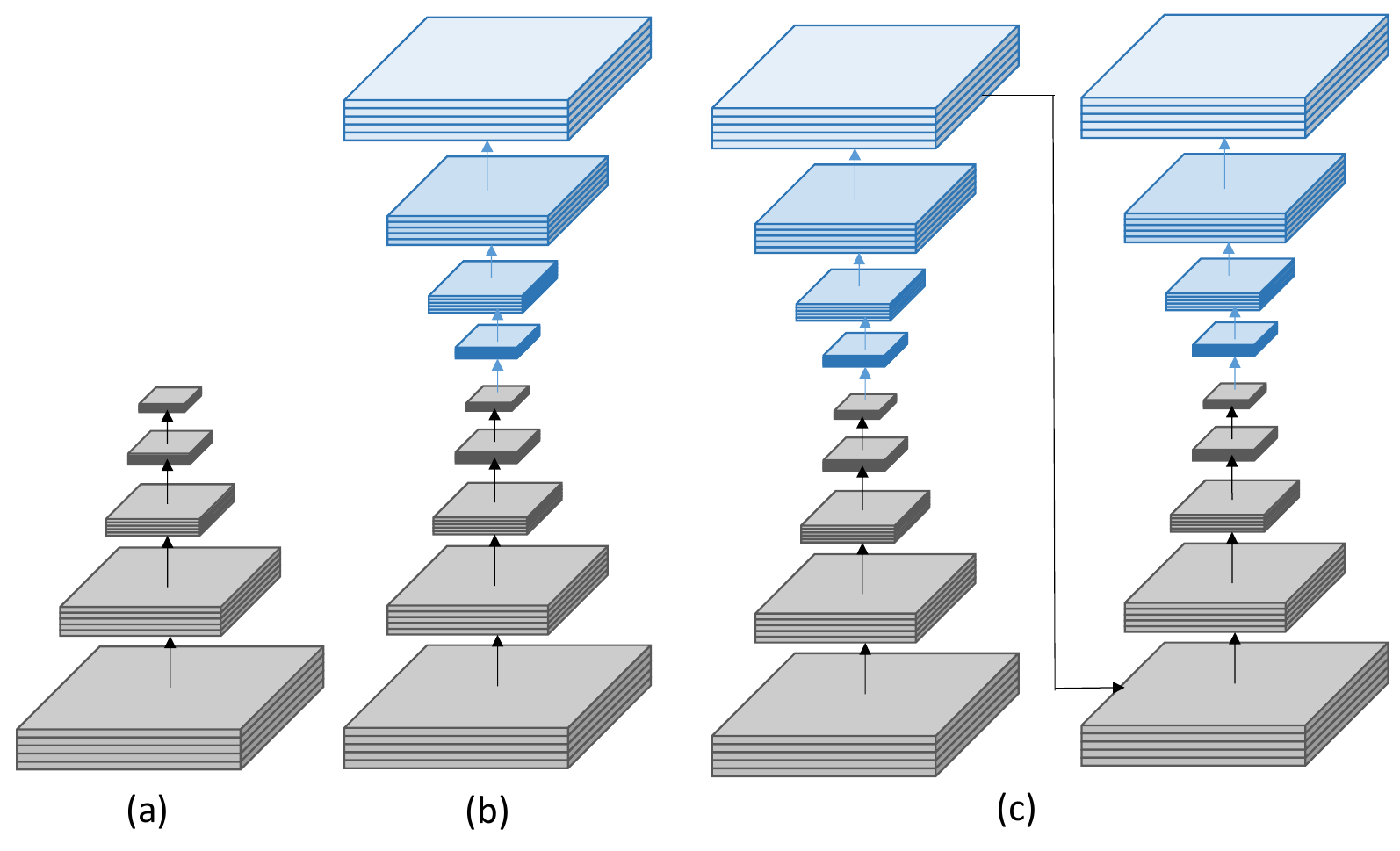} }}%
    \qquad
    \subfloat[VGG-16 with 5 convolutional layer and 5 deconvolutional layers on which DRLS is built]{{\includegraphics[width=0.45\columnwidth]{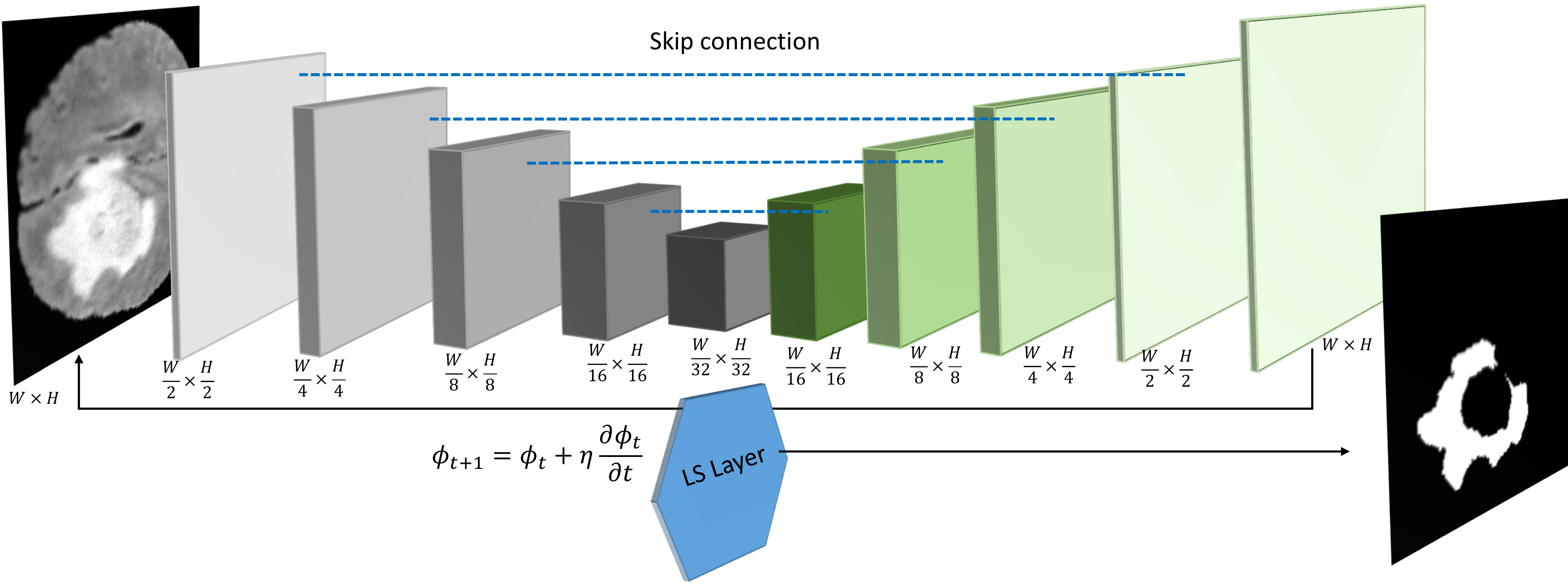} }}%
    \caption{The details of the proposed DRLS}%
    \label{fig:detail}%
\end{figure}

\cite{matan1992multi} extended the classic CNNs to infer and learn from arbitrary-sized inputs. Later, \cite{long2015fully} proposed a Fully Convolutional Neural Network (FCN) model which adapts and extends deep classification architectures to learn efficiently from whole input and whole ground truth images. By casting fully connected layers into convolutional neural network with kernels that cover their entire input regions, FCN allows to take input of any size and generate spatial outputs in one forward pass. To map the coarse feature map into a pixel-wise prediction of the input image, FCN up-samples the coarse feature map by a stack of deconvolution layers. Figures \ref{fig:detail} (a-1, a-2) show the comparison between classic CNN and FCN.

The recurrent fully-convolutional network (RFCN) is an extension of FCN architecture and given in \ref{fig:detail} (a-3). In the proposed RFCN, the output feature map of the current step is the input to the next step.

%In the proposed RFCN, the network initials with the input image and the initial contour $C$ and presented by LS function $\phi$

\subsection{Deep Recurrent Level Set (DRLS) - Proposed}
\label{subsec:RLS}
 To overcome the limitation of traditional LS, we incorporate LS into deep network that makes it more robust and powerful to deal with complex images. Our proposed DRLS is based on an observation that LS segmentation uses gradient descent which is seamless for deep framework to solve the brain tumor segmentation. Our proposed DRLS network is built based on VGG-16 with three different layers: convolutional layer, deconvolutional layer and LevelSet layer as shown in Figure \ref{fig:flowchart}.

\subsubsection{Convolutional Layer:}
Convolutional layers form the building blocks of a CNN. They have shared weights which are across different spatial locations and are defined on a translation invariance. The input and output of each convolutional layer is a feature map (tensor). The output feature map is computed by convolving the input feature map with convolution kernels $\textbf{Y}^{(s, \theta)} = f^s(\textbf{X}, \theta) = \textbf{X}*\textbf{W}^s + \textbf{b}$
% \begin{eqnarray}
% \begin{split}
% \textbf{Y}^{(s, \theta)} &= f^s(\textbf{X}, \theta) = \textbf{X}*\textbf{W}^s + \textbf{b}
% \end{split}
% \end{eqnarray}
where  $\textbf{X}$ is the input feature map. $\textbf{W},\textbf{b}$  are convolution kernel and bias.   $\textbf{W}^s$ indicates convolution at a stride $s$. $\textbf{Y}^{(s, \theta)}$ denotes the output feature map generated by the convolutional layers with total stride of $s$ and parameterized by $\theta$.  
%To improve translation invariance and visual representation ability, the convolutional layers are usually interleaved with max pooling layers and non-linear units ReLUs.  Different from CNN, the output feature map at the last convolutional layer is casted into convolutional layers with kernels that cover their entire input regions. As a result, the network allows to take an input image of arbitrary size. 
Because of the stride of convolutional and pooling layers, the final output feature maps $\textbf{Y}^{(s, \theta)}$ is downsampled by a factor of the total stride of $s$ compared to the input feature map. 
%To upsample the coarse output feature maps to same spatial size of the input feature maps, the network is designed with a stack of deconvolutional layers as shown in Fig.\ref{fig:proposed_DRLS} (a-b)
\subsubsection{Deconvolutional Layer:}
The deconvolutional layer is used to upsample the input feature maps using the stored max-pooling indices from the corresponding convolutional feature map. Here, a skip connection is introduced to concatenate the output of deconvolutional feature map with the corresponding convolutional feature map. Figure \ref{fig:detail} (b) illustrates the network's architecture.

A deconvolutional layer takes the output feature maps ($\textbf{Y}^{(s,\theta)}$) from the previous convolutional layer as its input feature maps. 
Let $g^s(; \tau)$ denote a deconvolutional layer parameterized by $\tau$ that up-samples  the input by a factor of $s$. The output is then concatenated with the corresponding convolutional layer $\textbf{Y}^{(s-1,\theta)}$ via skip connection as $\hat{\textbf{Y}}^{s, \tau} = concat\left[ g^s(\textbf{Y}^{(s,\theta)}; \tau), \textbf{Y}^{(s-1,\theta)}\right]$. 
% \begin{equation}
% \hat{\textbf{Y}}^{s, \tau} = concat\left[ g^s(\textbf{Y}^{(s,\theta)}; \tau), \textbf{Y}^{(s-1,\theta)}\right]
% \end{equation}
Unlike the simple bilinear interpolation, parameters $\tau$ of deconvolutional layers are jointly learned.  

%We denote $h$ as a function to extract visual representation of an input $\textbf{x}$ by the both convolutional layers and deconvolutional layers. The output $\textbf{y}$ is the deconvolutional feature map at the finest layer

% \begin{equation}
% \textbf{y} = h(\textbf{x})
% \end{equation}

\subsubsection{LevelSet Layer}
% To incorporate the RFCN with LS framework, we first convert the output feature maps (\textbf{y}) from FCN to [-0.5, 0.5] treat it as level set function $\phi$. This process is done by applying Euclidean distance transform ($\xi$) into $\textbf{y}$. At the first step, the proposed DRLS starts with the input image \textbf{I}
% \begin{eqnarray}
% \begin{split}
% \nonumber
% \textbf{y} &= h(\textbf{I})\\
% \phi_0 &= \xi(\textbf{y}) -  \xi(1 - \textbf{y}) + \textbf{y} - 0.5
% \end{split}
% \end{eqnarray}
 
To incorporate the RFCN with LS framework, the output feature maps are converted  (\textbf{y}) from FCN to [-0.5, 0.5] via Euclidean distance transformation ($\xi$) to treat it as a level set function $\phi = \xi(\textbf{y}) -  \xi(1 - \textbf{y})$
The proposed DRLS, refers to  the input space $\phi$ as $\Omega$. The network is trained to minimize the following energy function:
\begin{eqnarray}
\begin{split}
E(c_1, c_2, \phi) &= \int_{\Omega}\mu {H(\phi)} + \nu {\delta(\phi)|\nabla\phi| } \\
& + \alpha(H(\phi) - GT)^2+ \lambda_1 {|H(\phi)-c_1|^{2}H(\phi)  + \lambda_2 |H(\phi)-c_2|^{2}(1-H(\phi)) } dxdy 
\label{eq:energy} 
\end{split}
\end{eqnarray}
 In Eq.\ref{eq:energy}, the first term defines the area inside the contour C whereas the second term defined the length of the contour $C$ (segmentation boundary). The first term is ignored by setting $\mu  = 0$. Unlike the traditional VLS which sets $\nu > 0$ to be robust to noise, here $\nu$ is set as $\nu < 0$ to get more information regarding the different shapes of a brain tumor. In the third term, $GT$ is the groundtruth. Minimizing this term with $\alpha > 0$ supervises the network to learn where a brain tumor occurs in the MRI images. The last two terms correspond to energy inside and outside of the contour $C$. To force the feature map to be uniform on both inside and outside of the brain tumor regions, $\lambda1, \lambda2$ are set to be positive. $c_1$ and $c_2$ are two constants. 
To optimize the energy function, the calculus of variations is used. The derivative of energy function $E$ w.r.t $\phi$ is,
\begin{eqnarray}
\begin{split}
\frac{\partial E}{\partial \phi}  & = \delta(\varphi)\left [ \mu -\nu div\frac{\bigtriangledown\phi}{|\bigtriangledown\phi|} + 2\alpha (H(\phi)- GT) + \lambda_1(H(\phi)-c_1)^2\right. \\
& \quad \left. {}  + 2\lambda_1(H(\phi) - c_1)H(\phi) - \lambda_2(H(\phi) - c_2)^2 + 2\lambda_2(H(\phi) - c_2) (1-H(\phi)) \right ]
\end{split}
\end{eqnarray}
The derivatives of energy function $E$ w.r.t $c_1$ and $c_2$ are,
\begin{eqnarray}
\begin{split}
\frac{\partial E}{\partial c_1} &= -2\lambda_1(H(\phi) - c_1)H(\phi) \ \ \ \ 
\frac{\partial E}{\partial c_2} &= -2\lambda_2(H(\phi) - c_2)(1-H(\phi))
\end{split}
\end{eqnarray}
By maintaining a fixed $\phi$ and minimizing the energy function, $c_1$ and $c_2$  are calculated to be the average values of the inside and outside of the contour as,
\begin{equation}
%\small
\begin{split}
c_1  = \frac{\int_{\Omega}{H(\phi)(x,y) H(\phi) dxdy}}{\int_{\Omega} H(\phi)dxdy} \text{ , }
c_2 =  \frac{\int_{\Omega}{H(\phi)(x,y)(1-H(\phi))dxdy}}{\int_{\Omega}(1-H(\phi))dxdy}
\end{split}
\label{eq:c}
\end{equation}
Maintaining fixed $c_1$ and $c_2$, and minimizing the energy function w.r.t $\phi$, the associated Euler–Lagrange equation for $\phi$ is deduced. The descent direction is parameterized by a time $t>0$ and the level set $\phi$ is updated as:
\begin{eqnarray}
\begin{split}
%\phi_t & = \phi_{t-1} + \eta \frac{\partial \phi}{\partial t} \\
\frac{\partial \phi}{\partial t} & = \delta_\epsilon(\varphi)\left [ -\mu +\nu div\frac{\bigtriangledown\varphi}{|\bigtriangledown\varphi|}  -\lambda_1(H(\phi)-c_1)^2 - 2\lambda_1(H(\phi) - c_1)H(\phi) \right. \\
& \quad \left. {}  +\lambda_2(H(\phi) - c_2)^2 - 2\lambda_2(H(\phi) - c_2) (1-H(\phi)) \right ] 
\end{split}
\end{eqnarray}
Unlike the traditional deep learning methods, the predicted segmentation by the proposed method is computed via a Heaviside function $H(\phi)$ defined in Section \ref{sec:LSformular}. In the context of brain tumor segmentation, foreground measurement is of higher interest i.e. either flair or T1 or T2 or T1c of each pixel in the given image. At the initial stage, the DRLS takes an input image $\textbf{I}$ of $W\times H\times 1 $ as input feature maps and  generates the output feature maps $\textbf{x}$ of $W\times H\times 1 $. As shown in Figure \ref{fig:detail}(b), the proposed DRLS is built on VGG-16 with 5 convolutional layer and 5 deconvolutional layers. The  DRLS forward procedure is given in the Algorithm \ref{fig:flowchart}(b)
\begin{comment}
\begin{algorithm}[H]
\caption{DRLS learning procedure}
\begin{algorithmic} 
\State {\textbf{Input}: Image \textbf{I}}
\State {\textbf{Output}: Segmenting result \textbf{S}}
\State{\textbf{Procedure}}
\State{\# visual representation by FCN}
\State{$\textbf{y}$ = h($\textbf{I}$) }
\State{\# Convert to [-0.5, 0.5]}
\State{$\phi = \xi(\textbf{y}) -  \xi(1 - \textbf{y}) + \textbf{y} - 0.5$ }
\For{$i = 0$ to ${nsteps}$}
	\State{$\phi_{old} = \phi$}
    \State{\#LS curve evolution}
    \State{$\phi_{new} = \phi_{old} + \eta\frac{\partial \phi_{old}}{\partial t}$ } 
    \State{\# visual representation by FCN}
    \State {$\textbf{y} = h(\phi_{new})$ }
    \State{\# Convert to [-0.5, 0.5]}
    \State{$\phi= \xi(\textbf{y}) -  \xi(1 - \textbf{y}) + \textbf{y} - 0.5$ }
\EndFor
\label{al:1}
\end{algorithmic}
\end{algorithm}
\vspace{-1em}
\end{comment}
\subsubsection{Implementation Details}
 Figure \ref{fig:detail}(b) shows the structure of the proposed DRLS network which is built upon VGG16-based FCN. The proposed network is composed of three layers, i.e. convolutional layer, deconvolutional layer and LevelSet layer. ReLU layers are applied after each convolutional layers. LevelSet layer take $\phi_i$ produced by FCN network as an input. An initial learning rate of 1e-4 is used to update the weights and it is reduced when validation performance stops improving. The network is implemented over a PyTorch framework in Python environment and runs in a machine of Core i7-6700 @3.4GHz CPU, 64.00 GB RAM and a single NVIDIA GTX Titan X GPU. To setting up the training and testing, the BRATS 2017 dataset is divided into 80\% training set and 20\% testing set.

\section{Experimental Results}
%\subsection{Dataset \& Measurements}
% PLEASE GIVE SOME DESCRIPTION FOR THE BRATS DATASET
% ADD CIATAION 
% ADD THE FOLLOWING TABLE AND GIVE SOME DESCRIPTION ABOUT THE HGG & LGG

\textbf{Dataset \& Measurements}: The proposed DRLS method is evaluated on BRATS 2017 data. Additionally, BRATS 2013 and BRATS 2015 datasets were used for comparing its performance with the state of the art techniques. BRATS 2017 dataset is provided by MICCAI for automated brain tumor segmentation task. Each dataset contains two subsets corresponding to LGG (grade one and grad two) and HGG (high grade gliomas). The dataset is divided into training (80\%) and testing (20\%) datasets. The network is first trained on 168 HGG and 60 LGG
training set of BRATS 2017 and then is fine tuned on the training sets of BRATS 2015 and 2013, respectively. 

\textbf{Results:}
The algorithm was tested on 42 HGG and 15 LGG patients from BRATS 2017 data, 44 HGG and 11 LGG patients from BRATS 2015 and 9 HGG and 7 LGG patients from BRATS 2013 for comparison with other methods. The algorithm integrated merits on both level sets and deep learning and performs at par with the state of the art methods when run on local machines.
%The proposed algorithm was tested on 2017 BRATS dataset. 80\% of the patients were chosen as training samples while 20\% of the patients were chosen as testing samples. Thus 168 HGG images and 60 LGG images are used for training while, 42 HGG images and 15 LGG images are used for testing in the BRATS 2017 dataset. The 3D MRI images of each patient for all the 4 modalities were converted to slices of 2D images. 2D images corresponding to a modality were feed forward into the DRLS architecture to obtain segmentations for a particular region of tumor. The four modalities FLAIR, T2, T1-T1c are feed forward for each patient to obtain segmentations of tumor regions namely, Whole tumor, tumor core, enhancing tumor structures(necrotic/cystic core and enhancing core) respectively. Following evaluation metrics were observed:

\textbf{Evaluation metrics:} To evaluate the performance of the proposed method, standard metrics are used as suggested in BRATS challenge \cite{6975210} namely, Dice score, Sensitivity(Sens) score and Specificity (Spec). Besides metric scores, time consumption is also a key factor . Certain methods such as Tustison et al \cite{5} take 100 minutes to compute predictions per brain. However, when run on 4 GPUs the proposed algorithm shows a run time of just 55 seconds per patient.
Overall, on 2017 BRATS dataset, the DRLS algorithm achieved an average Dice score 0.86, 0.89 and 0.77 for Whole tumor(WT), Core Tumor(CT) and Enhancing core tumor(CT) regions respectively. Thus, achieving high performance and hence is on equal footing with the other state of the art methods. 
The sensitivity and specificity values achieved on BRATS 2017 dataset are (0.89, 0.88, 0.91) and (0.88, 0.78, 0.73), respectively.
\begin{comment}
summarized along with Dice score in Table \ref{Tab:2017}
\begin{table}[H]
\centering
\caption{Summarized Results obtained for proposed DRLS algorithm}
\label{Tab:2017}
\begin{tabular}{|l|l|l|l|l|}
\hline
Evaluation Metric  & Whole Tumor &  Tumor Core & Enhancing tumor \\ \hline
Dice Score &  0.86 & 0.89 & 0.77  \\ \hline
Sensitivity & 0.89 & 0.88 & 0.91  \\ \hline
Specificity & 0.88 & 0.78 & 0.73  \\ \hline
\end{tabular}
\end{table}
\end{comment}
%The algorithm was now subjected to testing with 2015 and 2013 BRATS datasets to analyze its performance further. 
Although, the algorithm can be compared using dice scores, the comparisions cannot be considered as completely fair since certain algorithms such as Urban et al \cite{9} and Chang et al\cite{13} do not report sensitivity  values. Additionally, most of the algorithms do not report specificity values. Furthermore, \cite{10} do not report results on BRATS 2013 test data-set. 
%However, all the techniques used for comparison implement deep learning and are similar to the proposed algorithm. Moreover, they are state of the art solutions for brain tumor segmentation and hence form good standards to compare against.
A summary of the comparison of the algorithm with BRATS 2013 and 2015 dataset are provided in Table \ref{Tab:2013} and \ref{Tab:2015}.
\begin{table}
\centering
\caption{Performance of DRLS in comparison with other methods when tested on BRATS 2013 dataset}
\label{Tab:2013}[H]
\begin{tabular}{|c|c|c|c|c|c|c|c|c|c|}
\hline
Methodology              & \multicolumn{3}{c|}{Dice Score} & \multicolumn{3}{c|}{Sensitivity} & \multicolumn{3}{c|}{Specificity} \\ \hline
                         & WT        & CT        & ET      & WT        & CT        & ET       & WT        & CT        & ET       \\ \hline
Havei et al\cite{8}    & 0.88      & 0.79      & 0.73    & 0.87      & 0.79      & 0.80     & 0.89      & 0.79      & 0.68     \\ \hline
Urban et al\cite{9}    & 0.87      & 0.77      & 0.73    & 0.92      & 0.79      & 0.70     & -         & -         & -        \\ \hline
Zikic et al\cite{10}   & 0.837     & 0.736     & 0.69    & -         & -         & -        & -         & -         & -        \\ \hline
Pereira et al\cite{11} & 0.88      & 0.83      & 0.77    & 0.89      & 0.83      & 0.81     & -         & -         & -        \\ \hline
Proposed Algorithm       & 0.89      & 0.79      & 0.74    & 0.90      & 0.89      & 0.93     & 0.91      & 0.82      & 0.73     \\ \hline
\end{tabular}
\end{table}

Table \ref{Tab:2013} and Table \ref{Tab:2015} have shown that the proposed algorithm outperforms other methods in terms of Dice scores and Sensitivity. The performance can be credited to the availability of additional training data from 2017 BRATS dataset that helped in fine tuning hyper-parameters of the algorithm. 
%Similarly performance of the algorithm is tested on 2015 BRATS dataset and comparisons are drawn as summarized in Table \ref{Tab:2015}. Table \ref{Tab:2015} shows that the proposed algorithm, DRLS, shows equal or better evaluation metrics than other methodologies mentioned. 
\begin{table}[H]
\centering
\caption{Performance of DRLS in comparison with other methods when tested on BRATS 2015 dataset}
\label{Tab:2015}
\begin{tabular}{|c|c|c|c|c|c|c|c|c|c|}
\hline
Methodology                        & \multicolumn{3}{c|}{Dice Score} & \multicolumn{3}{c|}{Sensitivity} & \multicolumn{3}{c|}{Specificity} \\ \hline
                                   & WT        & CT       & ET       & WT        & CT        & ET       & WT        & CT        & ET       \\ \hline
Pereira et al\cite{11}           & 0.78      & 0.65     & 0.7      & -         & -         & -        & -         & -         & -        \\ \hline
Pavel et al\cite{12}             & 0.83      & 0.75     & 0.77     & -         & -         & -        & -         & -         & -        \\ \hline
Chang et al\cite{13}             & 0.87      & 0.81     & 0.72     & -         & -         & -        & -         & -         & -        \\ \hline
Deep Medic\cite{KAMNITSAS201761} & 0.896     & 0.754    & 0.718    & 0.903     & 0.73      & 0.73     & -         & -         & -        \\ \hline
DMRes\cite{14}                   & 0.896     & 0.763    & 0.724    & 0.922     & 0.754     & 0.763    & -         & -         & -        \\ \hline
Proposed Algorithm                 & 0.88      & 0.82     & 0.73     & 0.91      & 0.76      & 0.78     & 0.90      & 0.81      & 0.71     \\ \hline
\end{tabular}
\end{table}
As mentioned above, the results reported for DRLS pertain to the metrics achieved by testing on local machines on 20\% of the BRATS datasets. The algorithm is robust to outliers, runs fast and consistently shows improved core tumor segmentation.

\section{Conclusion}
In this paper, a novel algorithm for automatic brain tumor segmentation method using deep recurrent level sets that integrates the advantages of both deep learning and level set is proposed and the current state of the art solutions were briefly introduced. The results obtained confirm that by integrating level sets and recurrent FCN architectures the proposed DRLS is a  cutting-edge solution. Additionally, DRLS improves the speed of segmenting brain tumors to a large extent and thus making it a practical solution.

{\small
\bibliographystyle{ieeetr}
\bibliography{refs}

\begin{thebibliography}{10}

\bibitem{osher1988fronts}
S.~Osher and J.~A. Sethian, ``Fronts propagating with curvature-dependent
  speed: algorithms based on hamilton-jacobi formulations,'' {\em Journal of
  computational physics}, vol.~79, no.~1, pp.~12--49, 1988.

\bibitem{15}
S.~Ho, E.~Bullitt, and G.~Gerig, ``Level-set evolution with region competition:
  Automatic 3-d segmentation of brain tumors,'' in {\em ICPR}, 2002.

\bibitem{16}
S.Taheri, S.H.Ong, and V.F.H.Chong, ``Level-set segmentation of brain tumors
  using a threshold-based speed function,'' {\em Image and Vision Computing},
  vol.~28, no.~1, pp.~26--37, 2010.

\bibitem{17}
K.~Thapaliya, J.-Y. Pyun, C.-S. Park, and G.-R. Kwon, ``Level set method with
  automatic selective local statistics for brain tumor segmentation in mr
  images,'' {\em Computerized Medical Imaging and Graphics}, vol.~37, no.~7,
  pp.~522 -- 537, 2013.

\bibitem{long2015fully}
J.~Long, E.~Shelhamer, and T.~Darrell, ``Fully convolutional networks for
  semantic segmentation,'' in {\em Proceedings of the IEEE conference on
  computer vision and pattern recognition}, pp.~3431--3440, 2015.

\bibitem{chan2001active}
T.~F. Chan and L.~A. Vese, ``Active contours without edges,'' {\em IEEE
  Transactions on Image Processing (TIP)}, vol.~10, no.~2, pp.~266--277, 2001.

\bibitem{simonyan2014very}
K.~Simonyan and A.~Zisserman, ``Very deep convolutional networks for
  large-scale image recognition,'' {\em arXiv preprint arXiv:1409.1556}, 2014.

\bibitem{1}
Anitha.V and Murugavalli.S, ``Brain tumor classification using two-tier
  classifier with adaptive segmentation technique,'' {\em IET Comput.Vis},
  vol.~10, no.~1, pp.~9--17, 2016.

\bibitem{2}
Dimah.D, Nidhal.B, and Hassan.M.F, ``Assessing the non-negative matrix
  factorization level set segmentation on the brats benchmark,'' in {\em
  Proceedings MICCAI-BRATS Workshop 2016}, 2016.

\bibitem{3}
Havaei.M, Larochelle.H, Poulin.P, and Jadoin.P.M, ``Within-brain classification
  for brain tumor segmentation,'' {\em Int J Cars}, vol.~11, pp.~777--788,
  2016.

\bibitem{4}
Bauer.S, Nolte.L.P, and Reyes.M, {\em Fully automatic segmentation of brain
  tumor images using support vector machine classification in combination with
  hierarchical conditional random field regularization}, pp.~354--361.
\newblock 2011.

\bibitem{5}
Tustison, Nicholas.J, Shrinidhi.K.L, Wintermark, Max, and D.~et~al, ``Optimal
  symmetric multimodal templates and concatenated random forests for supervised
  brain tumor segmentation (simplified) with antsr,'' {\em Neuroinformatics},
  vol.~13, pp.~209--225, Apr 2015.

\bibitem{8}
M.~Havaei, A.~Davy, D.~Warde-Farley, A.~Biard, A.~Courville, Y.~Bengio, C.~Pal,
  P.-M. Jodoin, and H.~Larochelle, ``Brain tumor segmentation with deep neural
  networks,'' {\em Medical image analysis}, vol.~35, pp.~18--31, 2017.

\bibitem{9}
G.~Urban, M.~Bendszus, F.~Hamprecht, and J.~Kleesiek, ``Multi-modal brain tumor
  segmentation using deep convolutional neural networks,'' {\em MICCAI BraTS
  (Brain Tumor Segmentation) Challenge. Proceedings, winning contribution},
  pp.~31--35, 2014.

\bibitem{10}
D.~Zikic, Y.~Ioannou, M.~Brown, and A.~Criminisi, ``Segmentation of brain tumor
  tissues with convolutional neural networks,'' {\em Proceedings MICCAI-BRATS},
  pp.~36--39, 2014.

\bibitem{11}
S.Pereira, A.Pinto, V.Alves, and C.A.Silva, ``Brain tumor segmentation using
  convolutional neural networks in mri images,'' {\em IEEE Transactions on
  Medical Imaging}, vol.~35, no.~5, pp.~1240--1251, 2016.

\bibitem{12}
P.~Dvo{\v{r}}{\'a}k and B.~Menze, ``Local structure prediction with
  convolutional neural networks for multimodal brain tumor segmentation,'' in
  {\em International MICCAI Workshop on Medical Computer Vision}, pp.~59--71,
  Springer, 2015.

\bibitem{14}
K.~Kamnitsas, C.~Ledig, V.~F. Newcombe, J.~P. Simpson, A.~D. Kane, D.~K. Menon,
  D.~Rueckert, and B.~Glocker, ``Efficient multi-scale 3d cnn with fully
  connected crf for accurate brain lesion segmentation,'' {\em Medical image
  analysis}, vol.~36, pp.~61--78, 2017.

\bibitem{KAMNITSAS201761}
``Efficient multi-scale 3d cnn with fully connected crf for accurate brain
  lesion segmentation,'' {\em Medical Image Analysis}, vol.~36, pp.~61 -- 78,
  2017.

\bibitem{matan1992multi}
O.~Matan, C.~J. Burges, Y.~LeCun, and J.~S. Denker, ``Multi-digit recognition
  using a space displacement neural network,'' in {\em Advances in neural
  information processing systems}, pp.~488--495, 1992.

\bibitem{6975210}
B.~H. Menze, A.~Jakab, Bauer, {\em et~al.}, ``The multimodal brain tumor image
  segmentation benchmark (brats),'' {\em IEEE transactions on medical imaging},
  vol.~34, no.~10, pp.~1993--2024, 2015.

\bibitem{13}
P.~D. Chang, ``Fully convolutional neural networks with hyperlocal features for
  brain tumor segmentation,'' in {\em Proceedings MICCAI-BRATS Workshop 2016},
  pp.~4--9, 2016.

\end{thebibliography}
}
\end{document}